\documentclass[runningheads]{llncs}
\usepackage{multirow}
\usepackage[table,xcdraw]{xcolor}
\usepackage{graphicx}

\begin{document}
\title{Yelp Reviews and Food Types: A Comparative Analysis of Ratings, Sentiments, and Topics}
\titlerunning{Yelp Reviews and Food Types}
\author{Wenyu Liao\inst{1, \dag}\and
Yiqing Shi\inst{1, \dag}\and
Yujia Hu\inst{1, 2}\and
Wei Quan\inst{1, 2}
}
\authorrunning{W. Liao et al.}
\institute{Department of Statistics and Data Science, BNU-HKBU United International College, Zhuhai 519087, China \\
\email{\{q030026082,q030026125\}@mail.uic.edu.cn}\\ 
\and
Guangdong Provincial Key Laboratory IRADS, BNU-HKBU United International College, Zhuhai 519087, China\\
\email{\{yujiahu,weiquan\}@uic.edu.cn}\\ 
\dag These authors contributed equally to this work and share first authorship
}

\maketitle              
\begin{abstract}
This study examines the relationship between Yelp reviews and food types, investigating how ratings, sentiments, and topics vary across different types of food. Specifically, we analyze how ratings and sentiments of reviews vary across food types, cluster food types based on ratings and sentiments, infer review topics using machine learning models, and compare topic distributions among different food types. Our analyses reveal that some food types have similar ratings, sentiments, and topics distributions, while others have distinct patterns. We identify four clusters of food types based on ratings and sentiments and find that reviewers tend to focus on different topics when reviewing certain food types. These findings have important implications for understanding user behavior and cultural influence on digital media platforms and promoting cross-cultural understanding and appreciation.

\end{abstract}

\section{Introduction}
Analyzing online reviews has become increasingly important over the past decade as more and more people rely on them to make decisions. Online reviews can provide valuable information about the quality, value, and overall experience of products and services from the perspective of other people. 

Yelp is one of the largest online review platforms, with millions of reviews and ratings for businesses across various industries. According to \cite{sara2017authenticity}, the large sample size of Yelp reviews reflects the dominant culture, which is continuously and historically rooted in favoring the white, Euro-centric experience. The average Yelp reviewer focuses on dirt floors, plastic stools, and other patrons who are non-white when reviewing non-European restaurants. However, the reviewers associate more positive characteristics when discussing European cuisines. 

Stereotypes and biases can be deeply ingrained in cultural attitudes toward food, and these attitudes may manifest in the way that people review and rate restaurants. Thus, this study investigates the relationship between Yelp reviews and food types, investigating how ratings, sentiments, and topics vary across different types of food in the Yelp dataset. Specifically, we analyze how ratings and sentiments of reviews vary across food types, cluster food types based on their ratings and sentiments, and compare topic distributions among different food types. We aim to answer the following research questions:
\begin{itemize}
\item1. What is the mean rating distribution of food types? Do reviewers tend to rate different food types differently?
\item2. What is the proportion of sentiment distribution of food types? Do reviewers tend to express different sentiments towards different food types?
\item3. What are the clusters of food types based on reviewer ratings and sentiments?
\item4. What are the topic distributions of food types? Do reviewers tend to focus on different topics when they review different food types?
\end{itemize}

Our analyses can lead to various potential contributions, including understanding user behavior and cultural influence on digital media platforms. Our findings highlight the importance of recognizing and appreciating the diversity of food cultures and promoting cross-cultural understanding and appreciation.

\section{Background}
There has been a significant increase in research on online reviews, driven by the growing importance of online reviews as a tool for decision-making. Researchers have focused on various aspects of online reviews, such as analyzing online tourism reviews to explore racism \cite{li2020racism}, exploring characteristics of reviews and reviewers shape review frequency and continuity \cite{chen2013investigation}, among others. 

Some of the previous work on Yelp data includes studying reviews, race, and gentrification \cite{zukin2017omnivore}, investigating the cultural impact on social commerce \cite{nakayama2019cultural}, identifying bias \cite{choi2021empirical}, evaluating reviews of hospitals and experiences of racism \cite{tong2021evaluation}, examining whether a cuisine is part of mainstream \cite{boch2021mainstream}, exploring authenticity \cite{le2021exploring,yu2023food}, investigating negative public sentiment during the COVID-19 pandemic \cite{labelle2022pandemic}, and comparing reviewing behaviors on Chinese and French restaurants \cite{quan2022implicit}.

While there is a growing body of research on online reviews and their impact on businesses and consumers, there is still a need for a more detailed analysis of Yelp reviews on different food types. The gap our study intends to bridge is the need for more specific insights into the relationship between Yelp reviews and food types. Our study aims to fill this gap by conducting a comparative analysis of ratings, sentiments, and topics of Yelp reviews in the US and providing specific insights into how reviews vary across food types. By doing so, our study can contribute to the growing body of research on online reviews and provide helpful information for businesses, policymakers, and researchers interested in understanding user behavior and cultural influence on digital media platforms.

\section{Methods}
\subsection{Data}
In this study, we focus on restaurants in the US. We analyze the Yelp Open Dataset\footnote[1]{https://www.yelp.com/dataset}, a subset of Yelp's businesses, reviews, and user data for personal, educational, and academic purposes. We identify restaurants in ten popular food types in the US using Yelp restaurant categories\footnote[2]{https://blog.yelp.com/businesses/yelp\_category\_list/}, see Table \ref{tab:table1}. Then, we remove duplicate restaurants (cross-reference in ten food types) and filter out only English reviews. Our final dataset includes 1,152,452 reviews on 7,141 restaurants. 

\begin{table}[ht]
\centering
\caption{Food types, number of reviews, and categories}
\label{tab:table1}
\begin{tabular}{|c|c|c|c|c|c|}
\hline
\rowcolor[HTML]{B0B3B2} 
\textbf{FoodType} & \textbf{Review\#} & \textbf{Categories} & \textbf{FoodType} & \textbf{Review\#} & \textbf{Categories} \\ \hline
American & 511,848 & \begin{tabular}[c]{@{}c@{}}American\\ American (New)\\ American (Traditional)\end{tabular} & \cellcolor[HTML]{FFFFFF}French & \cellcolor[HTML]{FFFFFF}18,811 & \begin{tabular}[c]{@{}c@{}}French\\ Mauritius\\ Reunion\end{tabular} \\ \hline
\cellcolor[HTML]{FFFFFF}Japanese & \cellcolor[HTML]{FFFFFF}125,260 & \begin{tabular}[c]{@{}c@{}}Japanese\\ Conveyor Belt Sushi\\ Izakaya\\ Japanese Curry\\ Ramen\\ Teppanyaki\end{tabular} & \cellcolor[HTML]{FFFFFF}Chinese & \cellcolor[HTML]{FFFFFF}76,929 & \begin{tabular}[c]{@{}c@{}}Chinese\\ Cantonese\\ Dim Sum\\ Hainan\\ Shanghainese\\ Szechuan\end{tabular} \\ \hline
Italian & \cellcolor[HTML]{FFFFFF}141,160 & \begin{tabular}[c]{@{}c@{}}Italian\\ Calabrian\\ Sardinian\\ Sicilian\\ Tuscan\end{tabular} & \cellcolor[HTML]{FFFFFF}Mexican & \cellcolor[HTML]{FFFFFF}166,486 & \begin{tabular}[c]{@{}c@{}}Mexican\\ Tacos\end{tabular} \\ \hline
Greek & 23,050 & Greek & \cellcolor[HTML]{FFFFFF}Korean & \cellcolor[HTML]{FFFFFF}24,691 & \cellcolor[HTML]{FFFFFF}Korean \\ \hline
Indian & 23,190 & Indian & Thai & 41,027 & Thai \\ \hline
\end{tabular}
\end{table}

\vspace{-10pt} 

\subsection{Measurements}
Our dataset contains restaurant-level measurements, including food type, review count, star rating, and review-level measurement of star rating. We create topic and sentiment measurements at the review level for analysis purposes.

\subsubsection{Sentiment}
We use Vader \cite{hutto2014vader} to infer sentiments of reviews with three categories, including negative, neutral, and positive. Vader is a popular tool for sentiment analysis that uses a lexicon-based approach to infer the sentiment of a given text. It is particularly useful for analyzing short texts like tweets and product reviews, where traditional machine learning approaches may not be feasible due to the limited training data. 

\subsubsection{Topic}
We adopt the framework from previous studies \cite{nakayama2019cultural,ha2010perceived,jeong2011restaurant,ryu2012influence} to classify reviews into four major topics: food quality, service, ambiance, and price fairness. We train classification models with annotated data from \cite{SajnaniYelp}, then apply these models to infer reviews topics. Each review can have one or multiple topics. Specifically, we apply Natural Language Toolkit (NLTK) \cite{loper2002nltk} to process Yelp reviews. We employ a combination of grid search and cross-validation to train various machine learning models for the topic classification task, including Logistic Regression, Decision Tree, Support Vector Machine, Random Forest, and Naive Bayes. We use precision, recall, and F1-score as evaluation metrics. Then, we apply the best models in terms of the F1-score to infer topics for our reviews. 

\subsection{Analyses}
\subsubsection{Kolmogorov-Smirnov test}
The Kolmogorov-Smirnov (KS) test is a non-parametric statistical test for whether a frequency distribution is identical to the theoretical distribution or whether two samples come from the same distribution. It is a test that does not require assumptions about the population distribution. Given the large amount of data we have, the KS test is a highly efficient way to analyze the distribution of our sample, especially when dealing with large sample sizes. To further improve the reliability and accuracy of the KS test, we use the Bootstrap method to generate multiple re-sampled datasets and simulate the variability of the sample distribution. Using Bootstrap re-sampling, we can obtain a more reliable estimate of the test statistic and calculate confidence intervals that reflect the precision of our estimate.

\subsubsection{K-means Clustering Analysis}
Clustering analysis helps us understand the similarities and differences in review ratings and sentiment scores among food types, enabling comparison of food types. We use the fviz\_nbclust function to determine the optimal number of clusters. The function evaluates the clustering quality by calculating the within-cluster sum of squares (WSS) for different numbers of clusters. We choose the optimal number of clusters by observing the elbow or inflection point in the WSS curve. This point indicates a significant decrease in the improvement of the WSS as the number of clusters increases. Thus selecting this number of clusters yields better clustering results.

\section{Results}
\subsection{RQ1. What is the mean rating distribution of food types? Do reviewers tend to rate different food types differently?} 

We use the Bootstrap KS test to compare whether the ratings of two food types come from the same distribution. The results suggest that all of the ratings of the food types are from different distributions, which implies that there are significant differences between the food types in terms of how they are rated. This finding is important because it indicates that the food types may have different characteristics or experiences that affect their perceptions or ratings. 

\begin{figure}[ht]
    \centering
    \includegraphics[width=0.9\textwidth]{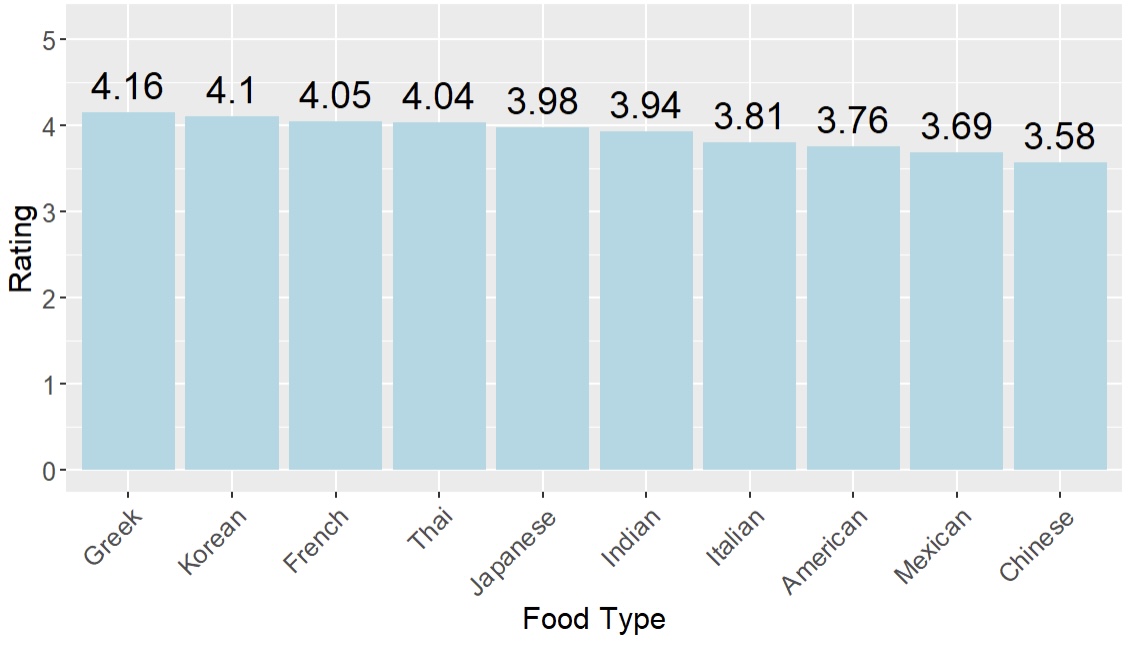}
    \caption{Distribution of Mean Ratings of Food Types}
    \label{rq1}
\end{figure}

Fig. \ref{rq1} shows that all ten food types have different mean ratings and that some are more likely to receive higher ratings than others. The food types arranged in descending order based on their mean ratings are Greek, Korean, French, Thai, Japanese, Indian, Italian, American, Mexican, and Chinese.

\vspace{-10pt} 

\subsection{RQ2. What is the proportion of sentiment distribution of food types? Do reviewers tend to express different sentiments towards different food types?}

\vspace{-10pt} 

\begin{figure}[ht]
    \centering
    \includegraphics[width=0.9\textwidth]{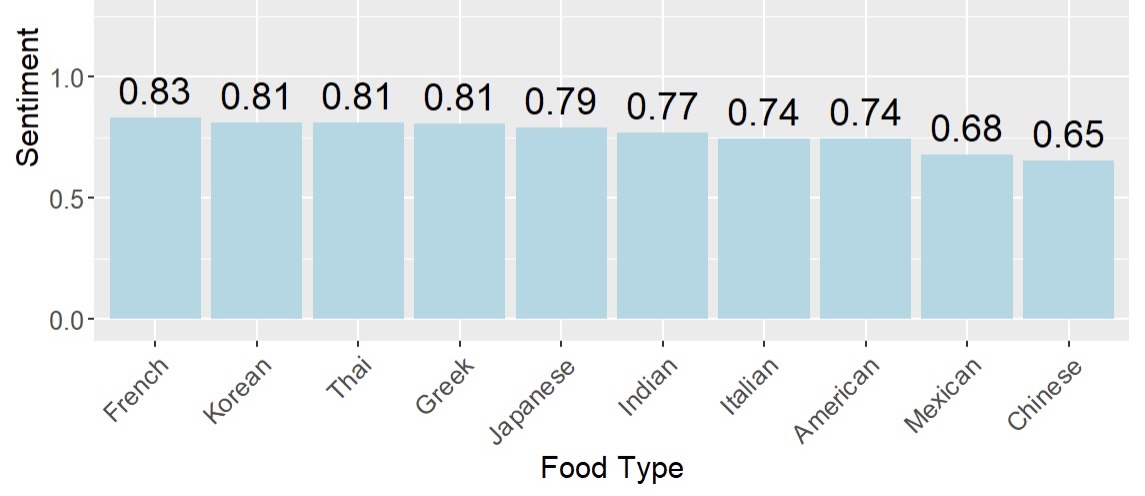}
    \caption{Distribution of Proportions of Sentiments of Food Types}
    \label{rq2}
\end{figure}

Similarly, we use the Bootstrap KS test to compare whether the sentiments of two food types come from the same distribution. The results show that America and Italian, Greek and Korean, Greek and Thai, and Korean and Thai have the same distribution. Other pairs of the sentiments of the food types are from different distributions, which implies that there are significant differences between the food types in terms of how reviewers express sentiments when they review.

Fig. \ref{rq2} shows that some food types have different proportions of sentiments and that some are more likely to receive positive sentiments than others. The food types arranged in descending order based on their average sentiment scores are French, followed by a group of Korean, Thai, and Greek, then Japanese, Indian, another group of Italian and American, Mexican, and finally, Chinese.

\subsection{RQ3. What are the clusters of food types based on reviewer ratings and sentiments?}

We use review star ratings and sentiments to cluster food types. We first standardize the data and conduct a normality test. We choose the optimal number of clusters to be four. Detailed results of food types and clusters are in Table \ref{tab:table2}. 

\vspace{-10pt} 

\begin{table}[ht]
\centering
\caption{Food types and Clusters}
\label{tab:table2}
\begin{tabular}{|l|l|l|l|l|}
\hline
\rowcolor[HTML]{B0B3B2} 
 & \multicolumn{1}{c|}{\cellcolor[HTML]{B0B3B2}\textbf{Cluster1}} & \multicolumn{1}{c|}{\cellcolor[HTML]{B0B3B2}\textbf{Cluster2}} & \multicolumn{1}{c|}{\cellcolor[HTML]{B0B3B2}\textbf{Cluster3}} & \multicolumn{1}{c|}{\cellcolor[HTML]{B0B3B2}\textbf{Cluster4}} \\ \hline
\multicolumn{1}{|c|}{\cellcolor[HTML]{D4D4D4}\textbf{Food Types}} & Chinese, Mexican & \begin{tabular}[c]{@{}l@{}}French, Greek,\\ Korean, Thai\end{tabular} & Indian, Japanese & American, Italian \\ \hline
\end{tabular}
\end{table}

\vspace{-10pt} 

\subsection{RQ4. What are the topic distributions of food types? Do reviewers tend to focus on different topics when they review different food types?}

\begin{figure}[ht]
    \centering
    \includegraphics[width=\textwidth]{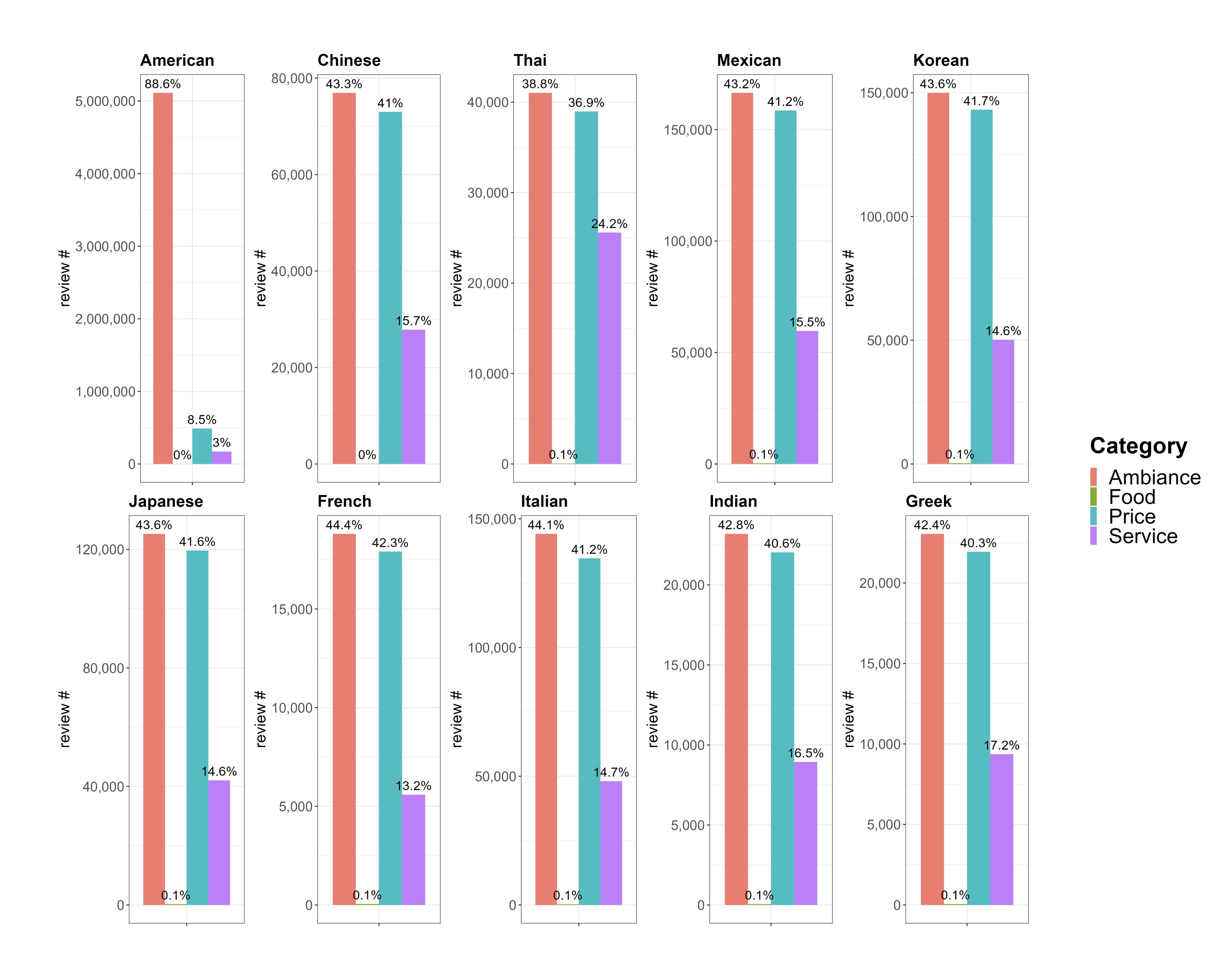}
    \caption{Distributions of Topics by Food Types}
    \label{rq4}
\end{figure}

We also use the Bootstrap KS test to compare whether the topics of two food types come from the same distribution. The results indicate that American, Italian, Japanese, and Korean have the same distribution; Indian and Thai have the same distribution; and Chinese and Mexican have the same distribution. The topics of other pairs of food types are from different distributions, which implies that there are significant differences between certain food types in terms of which topics reviewers focus on when they review. 

Fig. \ref{rq4}. shows the topic distributions by food types in all reviews. Reviewers tend to focus more on ambiance and less on food quality across all food types. Above 40\% of the reviews in Chinese, French, Greek, Indian, Italian, Japanese, Korean, and Mexican are about price fairness. Thai restaurants have less than 40\% of reviews on price fairness, and American restaurants only have 8.5\%. 24.2\% of reviews on Thai restaurants mention services, followed by Greek, Indian, Chinese, Mexican, Italian, Japanese, Korean, French, and American.

\section{Discussions}
The findings of RQ1 and RQ2 suggest significant differences in the ratings and sentiments of different food types. Various factors, including cultural factors, historical context, globalization, and health and nutrition concerns, could influence these differences. 

French cuisine has a long history of being associated with fine dining and upscale restaurants, which could lead to higher ratings due to perceptions of prestige and exclusivity. Similarly, Greek cuisine may be associated with healthy eating and the Mediterranean diet, which is popular in the US. Also, Greek and French cuisines may be more popular in urban areas of the US (where our Yelp reviews are from), with a higher concentration of upscale restaurants and foodies. This could lead to higher ratings due to a greater appreciation for these cuisines among restaurant reviewers. In the US, where there is a growing interest in healthy eating and wellness, these perceptions of health could contribute to higher ratings for Greek and French cuisines. 

Korean cuisine is becoming increasingly popular in the US, with more Americans seeking Korean food and flavors. This growing popularity could lead to higher ratings for Korean restaurants as more people become familiar with and appreciate Korean cuisine. Korean cuisine is known for its unique and bold flavors and use of ingredients that may be less common in other cuisines. This distinctiveness could lead to higher ratings due to the novelty and excitement of trying something new and different. Korean cuisine has become a popular topic on digital media platforms, with users sharing photos and videos of Korean dishes and restaurants. This exposure could lead to higher ratings for Korean restaurants as more people become interested in trying Korean food. 

Stereotypes and biases could play a role in the lower ratings of Mexican and Chinese restaurants. For example, Mexican cuisine may be associated with fast food or cheap, greasy fare, while Chinese cuisine may be associated with low-quality ingredients. These stereotypes could lead to negative perceptions and lower ratings. Also, there may be more competition among Mexican and Chinese restaurants in the US, leading to lower ratings due to higher expectations and more discerning diners. Additionally, the demand for these cuisines may be higher, leading to more casual or low-quality restaurants opening to meet demand and lower ratings overall.

Looking at Table \ref{tab:table2}, Chinese and Mexican cuisines may share some similarities in flavor profiles and ingredients, such as spicy or savory seasonings and sauces. French, Greek, Korean, and Thai cuisines cluster together. These cuisines may share some similarities in terms of their emphasis on fresh ingredients, bold flavors, and complex preparation techniques. Additionally, these cuisines may be perceived as more exotic or adventurous compared to other cuisines, which could lead to higher ratings and more positive sentiment among diners. Indian and Japanese cuisines are grouped together. These cuisines may share some similarities in terms of their use of spices and seasonings, their focus on fresh ingredients, and their simple preparation methods. Additionally, both cuisines have gained popularity in recent years, with more Americans seeking authentic Indian and Japanese food. American and Italian cuisines are in the same cluster. They may share similarities regarding their focus on comfort food and familiar flavors. Additionally, both cuisines are widely available in the US, which could lead to some overlap in the types of restaurants and dishes most commonly reviewed. It is also possible that the sentimental attachment that many Americans have to classic American dishes and Italian cuisine could contribute to higher ratings and more positive sentiments.

Findings from RQ4 reveal that some food types have the same topic distributions while others do not. This could be due to various factors, such as cultural and regional preferences, differences in the types of restaurants and dishes most commonly reviewed, and variations in the dining experience associated with different cuisines. For example, the similarities in topic distribution among American, Japanese, Korean, and Italian cuisines could be due to their popularity and widespread availability in the US, leading to more consistent patterns in the types of reviews and feedback reviewers provide. On the other hand, the differences in topic distribution among other food types could be due to their unique flavor profiles, preparation methods, and cultural associations, contributing to variations in the types of experiences and feedback reviewers have. Overall, these findings suggest significant differences in the topic distributions of different food types, and reviewers may focus on different topics when reviewing different types of cuisine. Further research would be needed to understand the reasons for these differences fully and identify the specific factors contributing to the topic distributions observed in this study.

\section{Conclusions}
In conclusion, our study has investigated the relationship between reviews and food types in the Yelp dataset. We have identified patterns and trends that shed light on reviewers' preferences and behavior through our analyses of rating and sentiment distributions, clusterings of food types, and topic distributions. Our findings suggest that different food types have distinct rating distributions and that reviewers tend to focus on different topics when reviewing specific types of food. Our study demonstrates the value of advanced analytical techniques for analyzing user-generated content. 

While our study provides valuable insights, it is important to consider these limitations when interpreting the results. Our study relies on the Yelp dataset, which may only represent some US restaurants. Yelp users may be a self-selected group of reviewers who are more likely to write reviews, and the types of restaurants and food types most commonly reviewed may differ from the restaurant industry as a whole. Additionally, the Yelp dataset may be subject to biases and inaccuracies, such as fake reviews or reviews that do not reflect actual dining experiences. Our study focuses on a limited set of variables, such as rating, sentiment, and topic distributions. While these variables are informative, they may not provide a complete picture of reviewers' preferences and behavior. Additionally, our study does not provide contextual information about the restaurants. For example, it does not account for differences in restaurant quality, the dining experience, or cultural and regional preferences that may influence user ratings and sentiments. Further research that addresses these limitations could provide a more comprehensive understanding of user preferences and consumer behavior.

\section*{Acknowledgment}
This work is supported in part by the Guangdong Provincial Key Laboratory IRADS (2022B1212010006, R0400001-22) and UIC Research Grant with No. of UICR0700038-22 at BNU-HKBU United International College, Zhuhai, PR China.

\bibliographystyle{splncs04}
\bibliography{resources/citation}

\end{document}